\documentclass{article}
\usepackage{ijcai25}

\usepackage{times}
\usepackage{soul}
\usepackage{url}
\usepackage[hidelinks]{hyperref}
\usepackage[utf8]{inputenc}
\usepackage[small]{caption}
\usepackage{graphicx}
\usepackage{amsmath}
\usepackage{amsthm}
\usepackage{booktabs}
\usepackage{algorithm}
\usepackage{algorithmic}
\usepackage[switch]{lineno}
\usepackage{lipsum}
\usepackage{amssymb}
\usepackage{xcolor}
\usepackage{mathrsfs}
\usepackage{fontawesome5}
\usepackage{wasysym}

\definecolor{golden}{RGB}{255, 215, 0}

\title{GarmentD\textcolor{golden}{\faIcon{bolt}}ffusion: 3D Garment Sewing Pattern Generation with Multimodal Diffusion Transformers}

\author{
Xinyu Li$^{1,2}$\footnote{Work done as a research intern at Shenfu Research.}
\and
Qi Yao$^{2}$\footnote{Project lead.}
\and
Yuanda Wang$^{2}$\\
\affiliations
$^1$Zhejiang University\\
$^2$Shenfu Research\\
\emails
lixinyu0801@zju.edu.cn,
\{yaoqi, adamwang\}@dejaai.com
}

\begin{document}

\maketitle

\begin{abstract}
Garment sewing patterns are fundamental design elements that bridge the gap between design concepts and practical manufacturing. The generative modeling of sewing patterns is crucial for creating diversified garments. However, existing approaches are limited either by reliance on a single input modality or by suboptimal generation efficiency. In this work, we present \textbf{\textit{GarmentDiffusion}}, a new generative model capable of producing centimeter-precise, vectorized 3D sewing patterns from multimodal inputs (text, image, and incomplete sewing pattern). Our method efficiently encodes 3D sewing pattern parameters into compact edge token representations, achieving a sequence length that is $\textbf{10}\times$ shorter than that of the autoregressive SewingGPT in DressCode. By employing a diffusion transformer, we simultaneously denoise all edge tokens along the temporal axis, while maintaining a constant number of denoising steps regardless of dataset-specific edge and panel statistics. With all combination of designs of our model, the sewing pattern generation speed is accelerated by  $\textbf{100}\times$ compared to SewingGPT. We achieve new state-of-the-art results on DressCodeData, as well as on the largest sewing pattern dataset, namely GarmentCodeData. The project website is available at \textit{\url{https://shenfu-research.github.io/Garment-Diffusion/}}.

\end{abstract}

\section{Introduction}

\begin{figure}[t]
    \centering
    \includegraphics[width=\linewidth]{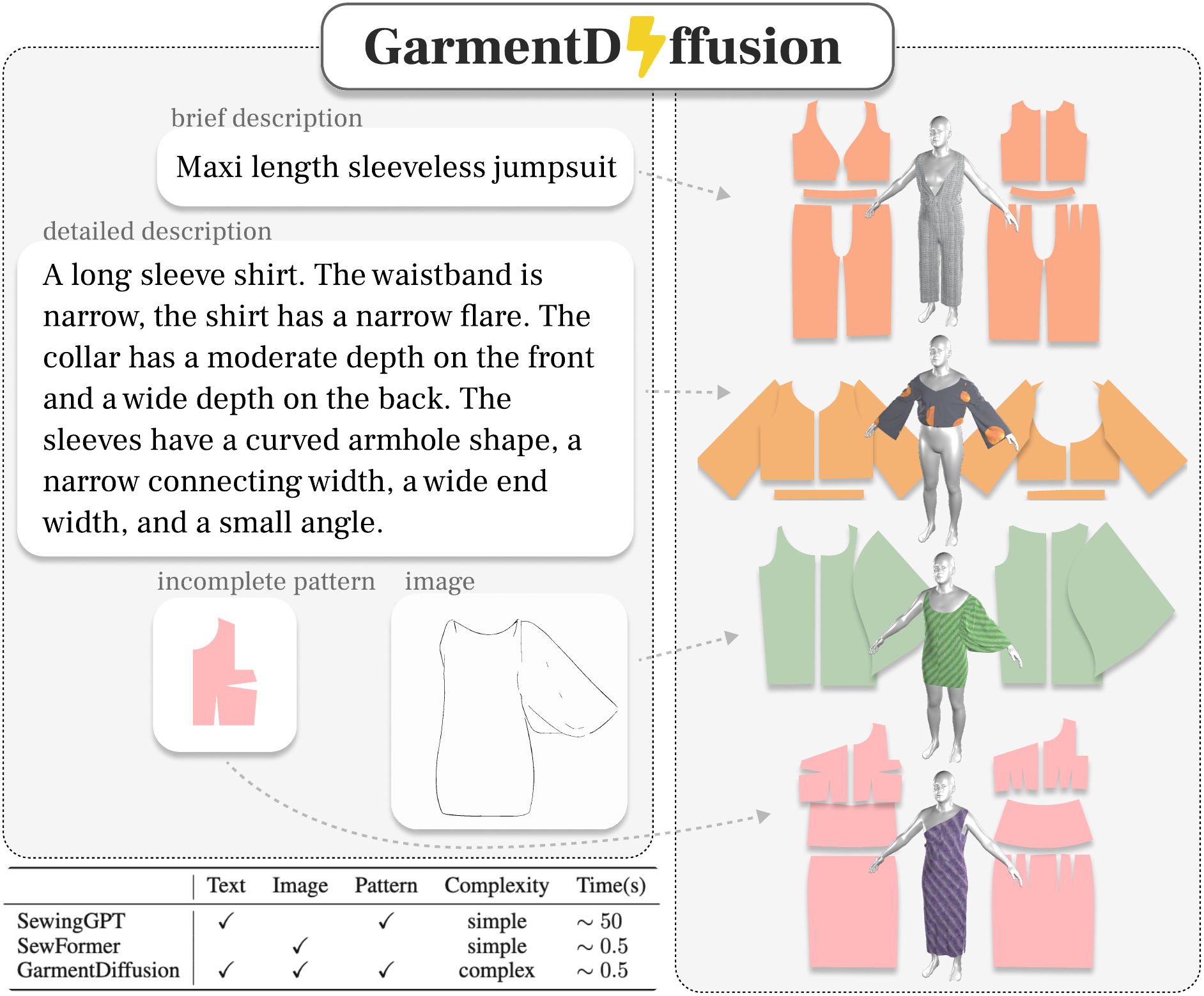}
    \caption{\textbf{3D garment pattern generation with multimodal inputs.} 
    As illustrated on the left, our model supports various input modalities, including brief and detailed text descriptions, images (garment sketches), and incomplete patterns. The generated patterns can be draped on human models and utilized for practical production. Our model supports both simple and complex pattern generation, e.g., DressCodeData and GarmentCodeData. As indicated by the \textbf{“lightning”} icon, our model demonstrates an ultra-fast pattern generation speed (within a second using a single A10 GPU), which is comparable to the discriminative model (SewFormer).
    }
    \label{fig:title}
\end{figure}

Digital garment modeling has emerged as a pivotal research area for fashion design, 
with garment sewing patterns being essential in transforming design concepts into tangible garments.
Many studies have been conducted on sewing pattern modeling and generation, 
either by scaling up the size of datasets~\cite{korosteleva2021generatingdatasets3dgarments,korosteleva2024garmentcodedatadataset3dmadetomeasure} 
or by proposing new parametric and learning-based approaches~\cite{korosteleva2022neuraltailor,Korosteleva_2023,Liu_2023,Chen_2024_WACV,he2024dresscode}.

The first attempt~\cite{korosteleva2021generatingdatasets3dgarments} was made to build a synthetic garment sewing pattern dataset using a parametric approach, with \textasciitilde20K sewing patterns. However, both the complexity of sewing pattern geometries and the quantity of sewing patterns are insufficient to meet the increasing data demands of advanced data-driven models. Subsequently, in~\cite{korosteleva2024garmentcodedatadataset3dmadetomeasure}, the authors built a large-scale garment dataset, i.e., GarmentCodeData, using component-oriented garment programs~\cite{Korosteleva_2023}. It introduces more complex design prototypes and scales up the number of sewing patterns by $5\times$, with \textasciitilde115K sewing patterns in total. However, learning-based models typically require paired samples for conditional training, such as {\tt (sewing pattern, modality-specific input)}. The unimodal nature of these datasets restricts the development of generative modeling approaches. SewFactory~\cite{Liu_2023} has recognized this problem and provides approximately 1M 
image-and-sewing-pattern pairs for training. Since its design prototypes are derived from~\cite{korosteleva2021generatingdatasets3dgarments}, the complexity of sewing pattern geometry is still limited.

Another challenge lies in efficiently modeling the generation of sewing patterns to make it more applicable to real-time scenarios. 
SewFormer~\cite{Liu_2023} introduces a DETR-like~\cite{carion2020end} discriminative model
to map 2D images to sewing patterns. The discriminative training of SewFormer leads to the \emph{deterministic} predictions of sewing patterns given input images, which restricts the diversity of garment designs compared to the generative approaches. A pioneering work, DressCode~\cite{he2024dresscode}, introduces a GPT-like autoregressive model (SewingGPT) to generate vector-quantized sewing patterns, conditioned on the text descriptions via cross-attention. While this method is effective in generating simple sewing patterns, 
it faces significant challenges when applied to GarmentCodeData~\cite{korosteleva2024garmentcodedatadataset3dmadetomeasure}. For example, the token sequence length of SewingGPT increases from \textasciitilde2K to over 18K, making its training and inference infeasible in practice. Another issue is the coarse sewing pattern descriptions generated by GPT-4V using its data annotation pipeline, which lacks the precise control for text-conditioned sewing pattern generation.

In this paper, we rethink the modeling paradigm of sewing patterns, questioning whether the vector-quantized encoding scheme and autoregressive \textit{next-parameter prediction} in DressCode are efficient for sewing pattern generation. Inspired by BRepGen~\cite{xu2024brepgen}, we encode edge-related parameters 
(such as 3D coordinates, stitch tags, and free edge scores) into the \textbf{embedding dimension}, 
while denoising all edge tokens in parallel along the temporal axis. By leveraging the parallel processing nature of diffusion transformers~\cite{peebles2023scalable}, our approach accelerates the generation process by approximately a hundredfold  
without sacrificing the precision (in centimeters) of sewing pattern geometries. 
Specifically, with parameters set to $\texttt{\#edge parameters/edge = 9}$ (endpoints, control points, arc), $\texttt{\#edges/panel = 39}$, and $\texttt{\#panels/pattern = 37}$,
SewingGPT requires $18,135 + 2 \text{(SOS,EOS)}$ tokens and steps to autoregressively generate a sewing pattern. 
In contrast, our model only needs 1,443 tokens for generation, with a constant denoising step independent of dataset statistics. Furthermore, following~\cite{khan2024text2cad}, we redesign the data annotation pipeline for both DressCodeData~\cite{korosteleva2021generatingdatasets3dgarments,he2024dresscode} and GarmentCodeData, to provide both brief and detailed text descriptions for sewing patterns. To support the image modality as input, we employ commonly used garment sketches as the interactive interface between users and models. As a benefit of our modeling paradigm, we also support sewing pattern completion using user-provided incomplete patterns as input for controllable generation.

To sum up, our contributions to the community are as follows:
\begin{enumerate}
    \item We present a new generative model, \textbf{\textit{GarmentDiffusion}}, pushing the limits of diffusion-based modeling paradigm for multimodal sewing pattern generation.
    \item We propose an efficient edge encoding scheme that significantly reduces the token sequence length of sewing patterns, achieving a substantial speedup compared with the autoregressive approach, i.e., DressCode.
    \item We validate the effectiveness of our model on SewFactory, DressCodeData, as well as the largest and most challenging GarmentCodeData, and establish a strong baseline with comprehensive and quantitative evaluation metrics.
    \item We provide new multimodal data annotation pipelines that can generate both brief and detailed text descriptions, as well as garment sketches for sewing patterns, enabling multimodal sewing pattern generation. 
\end{enumerate}

\section{Related Work}

\subsection{Garment Sewing Pattern Generation}
Existing research on garment generation can mainly be divided into 3D-based and sewing pattern-based approaches.
The 3D-based methods generates garment models
through Gaussian splatting guidance~\cite{li2024garmentdreamer3dgsguidedgarment}, 
unsigned distance function regression~\cite{moon20223d,Zheng_2024_CVPR}, 
neural volumetric rendering~\cite{chen2024single}, 
or latent representation learning~\cite{Su_2023,shen2020garmentgeneration,srivastava2025wordrobe,shao20243d}. 
However, these garment models often have topological imperfections that make them unsuitable for manufacturing.

To generate production-ready sewing patterns, early methods included 
scanned model flattening~\cite{bang2021estimating} and iterative panel parameter optimization~\cite{10.1145/3272127.3275074}. 
With the release of large sewing pattern datasets~\cite{korosteleva2021generatingdatasets3dgarments,korosteleva2024garmentcodedatadataset3dmadetomeasure}, there has been a shift toward data-driven approaches. 
SPnet~\cite{lim2024spnet} predicts sewing patterns from generated T-pose images, while Neural Sewing Machine~\cite{chen2022structure} utilizes principal component analysis to create sewing pattern masks. 
To reduce reliance on templates, 
Korosteleva and Lee~\shortcite{korosteleva2022neuraltailor} 
predict edges and stitches directly from 3D point clouds. 
Liu et al.~\shortcite{Liu_2023} and Chen et al.~\shortcite{Chen_2024_WACV} 
use hierarchical transformers to recover sewing patterns from images. 
He et al.~\shortcite{he2024dresscode} 
generate garment sewing patterns with textures, guided by natural language descriptions.
At the same time as our work, 
Design2GarmentCode~\cite{zhou2024design2garmentcodeturningdesignconcepts} proposes a DSL-oriented multimodal agent and leverage the garment programs to generate sewing patterns. 
ChatGarment~\cite{bian2024chatgarmentgarmentestimationgeneration} fine-tunes a VLM to generate garment specification files. AIpparel~\cite{nakayama2024aipparellargemultimodalgenerative} builds on top of LLaVA-1.5~\cite{liu2024improvedbaselinesvisualinstruction} and proposes to train a multimodal pattern generation model using both discrete and continuous training objectives. 
SewingLDM~\cite{liu2024multimodallatentdiffusionmodel} applies a latent diffusion model with two-stage training to generate patterns, incorporating various handcrafted losses. 
Our model adopts a single-stage training with MSE loss, achieving improvements in modality diversity, performance, and efficiency.

\subsection{Conditional Diffusion Models}
A large portion of current research on diffusion models originates from \cite{ho2020denoising,rombach2022high}, 
with a forward chain that perturbs data into noise and a reverse chain that converts noise back into data.
Diffusion Transformers (DiT)~\cite{peebles2023scalable} explore replacing the U-Net backbone 
with a transformer that operates on latent patches, achieving better scalability.
IP-Adapter~\cite{ye2023ip} introduces the decoupled cross-attention mechanism 
to achieve the image-prompt conditional generation.

Diffusion models have also proven successful in 
generating CAD models.
For example, Xu et al.~\shortcite{xu2024brepgen} present a diffusion-based approach that
unconditionally outputs boundary representation of CAD models. Wang et al.~\shortcite{WANG2024102327} utilize a vector-quantized diffusion model
to generate command sequences from design concepts represented by text or sketches.
The success of diffusion models in synthesizing structured CAD models suggests that 
similarly structured sewing patterns could also be generated through a reverse diffusion process.

\section{Method}
\subsection{Sewing Pattern Representation}
\label{sec:sewing pattern representation}
\begin{figure}[t]
    \centering
    \includegraphics[width=\linewidth]{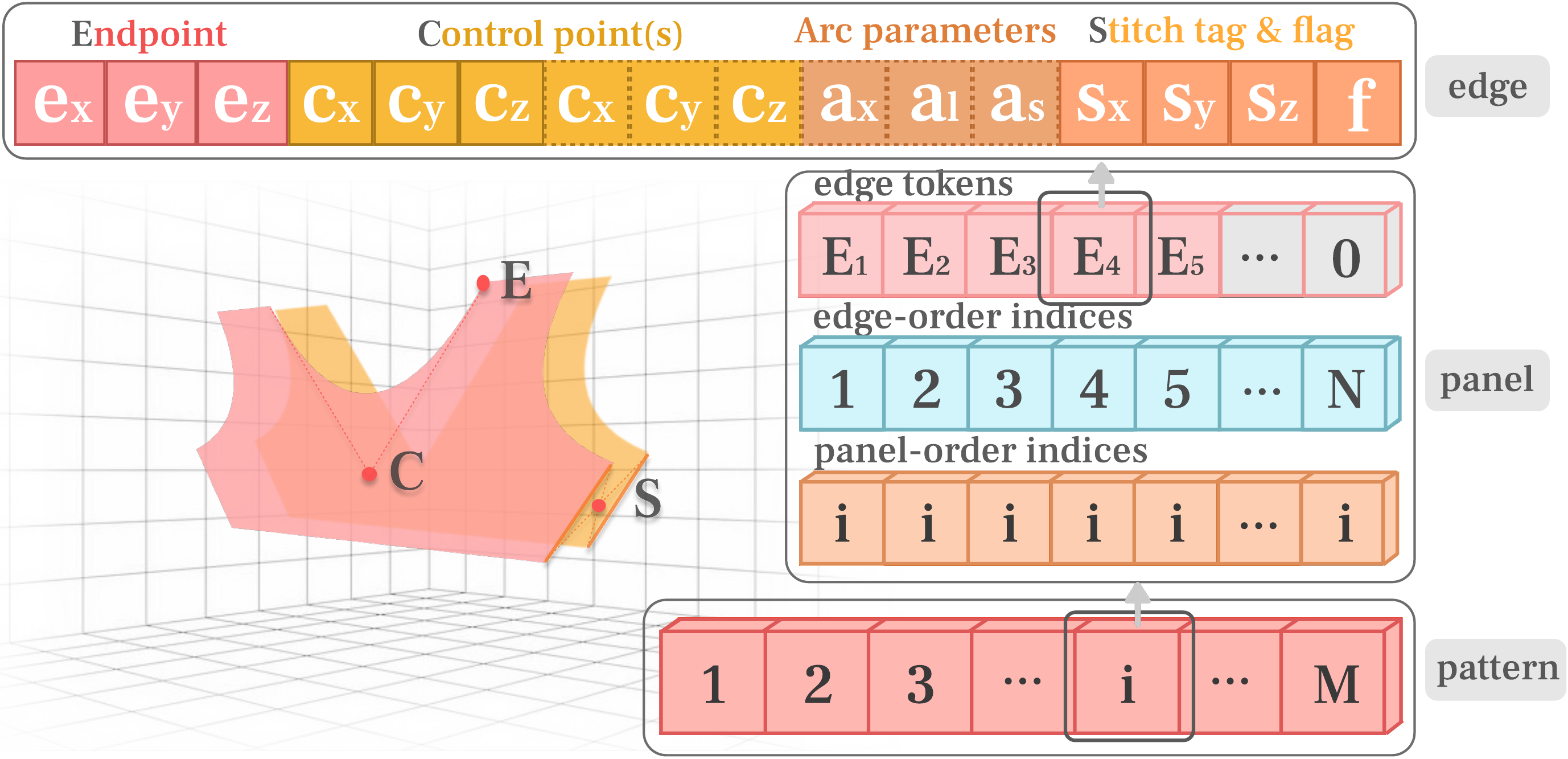}
    \caption{\textbf{Token representations for the edge, panel and pattern.} GarmentDiffusion utilizes an edge-oriented compact representation to encode the sewing pattern. After applying rotation and translation transformations to the 2D panels, the edge parameters are encoded along the embedding dimension. Each edge token is assigned an edge-order index and a panel-order index to indicate its global position within the sequence. The sequence is padded with zero tokens to ensure uniform length.}

    \label{fig:tokenize}
\end{figure}

\begin{figure*}
    \centering
    \includegraphics[width=\textwidth]{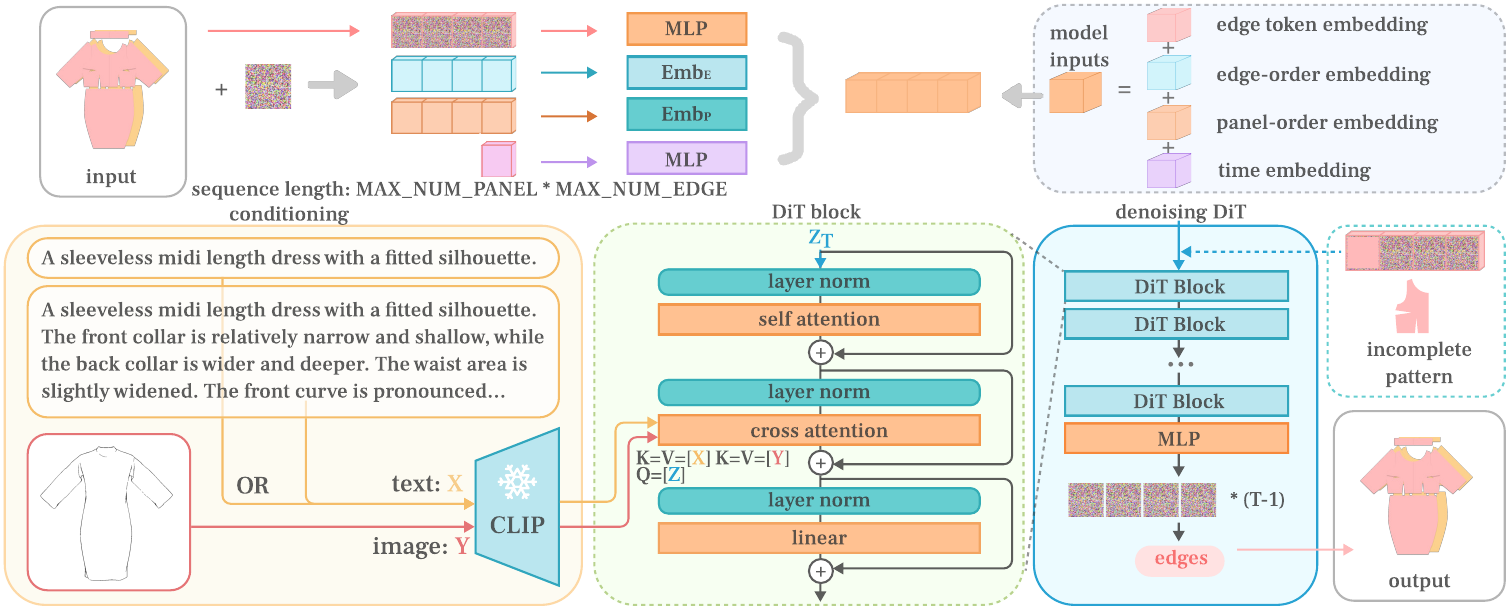}
    \caption{\textbf{The framework of GarmentDiffusion.} 
    GarmentDiffusion accepts multi-level text descriptions, a garment sketch or an incomplete sewing pattern as input conditions, and generates a sewing pattern through the denoising of random Gaussian noise. 
    The text and image features are extracted using a frozen CLIP and injected via decoupled cross-attention layers in each DiT block. The incomplete pattern replaces the initial subsequence of the random noise sequence for controllable generation. The final output is a 3D-placed sewing pattern that is consistent with the multimodal conditions.
    }
    \label{fig:network}
\end{figure*}

\begin{figure}
    \centering
    \includegraphics[width=\linewidth]{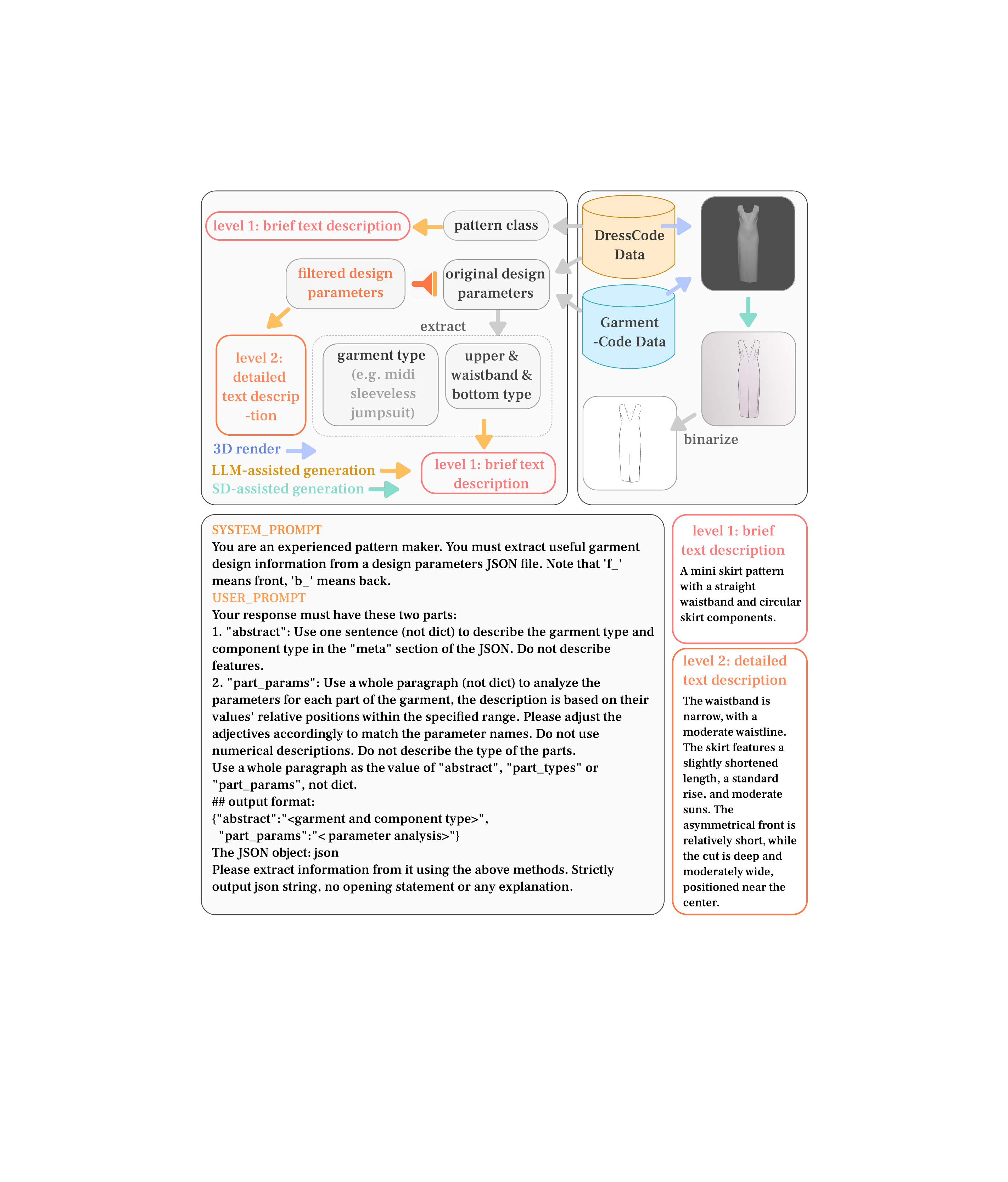}
    \caption{\textbf{Our multimodal data annotation pipelines.}
    The up-left pipeline illustrates the generation of text descriptions at both brief and detailed levels, 
    and the up-right pipeline represents the generation of garment sketches. 
    The system and user prompts for LLM and text descriptions examples are shown at the bottom.
    }
    \label{fig:data}
\end{figure}

\paragraph{Edge representation.}
A 3D pattern consists of variable number of panels, with each panel being a closed shape made up of multiple edges. Since the edges are connected end-to-end, we only need a start point $\mathbf{e}_{j} \in \mathbb{R}^3$ and control points $\mathbf{c}_{j} \in \mathbb{R}^3$ of Bézier curves to represent the geometry of $j^{\textit{th}}$ edge ${E}_j$, where $\mathbf{e}_{j} := (e_x, e_y, e_z)_j$ and $\mathbf{c}_{j} := (c_x, c_y, c_z)_j$ are the 3D point coordinates. The number of control points depends on whether the Bézier curve is quadratic or cubic. We represent the circular arc using another three parameters $\mathbf{a}_j \in \mathbb{R}^3$, where $\mathbf{a}_j := (a_x, a_l, a_s)_j$ represents the radius, major or minor arc and sweeping orientation, respectively. Note that if ${E}_j$ is a linear curve, $\mathbf{c}_{j}$ is identical to $\mathbf{e}_{j}$. Furthermore, we use a per-edge stitch tag $\mathbf{s}_{j} \in \mathbb{R}^3$ and a binary stitch flag $f_{j} \in \{0, 1\}$ to encode the stitch information of edges. $\mathbf{s}_{j} := (s_x, s_y, s_z)_j$ is calculated as the averaged 3D midpoint between matched edge
pairs. All coordinates are calculated after performing 3D rotation and translation for each panel.~${E}_j$ is thus represented as $\mathbf{e}_j \oplus \mathbf{c}_j \oplus \mathbf{a}_j \oplus \mathbf{s}_j \oplus f_{j}$ with the appropriate zero padding, where $\oplus$ represents the concatenation along the embedding dimension. Note that the number of parameters of ${E}_j$, denoted as $|{E}_j|$, could be variable length depending on the dataset.

\paragraph{Pattern representation.}
Suppose that a dataset contains at most $M$ panels for all patterns, and each panel contains at most $N$ edges. 
We pad the pattern $\mathbf{P} \in \mathbb{R}^{|m \times n| \times |E_j|}$ that has $m$ panels ($m \leq M$) and $n$ edges per panel ($n \leq N$) to the uniform $M \times N$ sequence length, which is denoted as $\mathbf{P}' \in \mathbb{R}^{|M \times N| \times |E_j|}$. That is, all panels $\{ P_i \}_{i > m}^M$ are set to $\mathbf{0}$, and all edges $\{ E_j \}_{j > n}^{N}$ are set to $\mathbf{0}$ as well, where $i$ and $j$ denote the index of panel and edge, respectively. The hierarchical pattern representation is illustrated in Figure~\ref{fig:tokenize}. This edge-oriented pattern representation largely shortens the token sequence length compared to the coordinate-oriented representation in DressCode. Different from the “token-by-token” causal generation, all these tokens can be processed by subsequent diffusion transformers \textbf{in parallel}.

\subsection{Pattern Generation with GarmentDiffusion}
Our model follows the design of DiT~\cite{peebles2023scalable} architecture. It accepts multimodal inputs to control the generation of sewing patterns. 
During the training phase, all edges are converted into token representations, followed by random panel shuffling and noise corruption. 
The model is trained to predict the noises added to the edge tokens. 
In the generation phase, the edge tokens are initialized as random Gaussian noises, and are iteratively denoised using the predicted noise.
The entire framework is illustrated in Figure~\ref{fig:network}.

\paragraph{Pattern preprocessing.}
Suppose we have $\mathbf{P}' \in \mathbb{R}^ {|M \times N| \times |E_j|}$, which consists of $\{ P_i \}_{i=1}^M$ with $\{ P_i \}_{i > m}^M$ being panel-level padding. Each $P_i$ consists of $\{ E_j \}_{j=1}^{N}$ with $\{ E_j \}_{j>n}^{N}$ being edge-level padding. We shift and scale each dimension of ${E}_j$ using respective parameter statistics to ensure that the value range is between -1 and 1. To support the pattern completion, we randomly shuffle $\{ P_i \}_{i=1}^m$ to break the predefined panel order, while keeping the order of edges $\{ E_j \}_{j=1}^{n}$ within a panel unchanged. We always place the padding tokens after the shuffled edge tokens.

\paragraph{Token embeddings.}
To distinguish the edge tokens among different panels, we construct a panel-level look-up embedding table $\mathbf{Emb}_P \in \mathbb{R}^{M \times C}$, where $C$ is the embedding dimension of the model. We also construct an edge-level look-up embedding table $\mathbf{Emb}_E \in \mathbb{R}^{N \times C}$ to encode the sequential edge order within a panel. To represent each time step $t$, we use the conventional sine and cosine positional encoding to construct the time embedding $\mathbf{t} \in \mathbb{R}^C$. We employ a DDPM noise scheduler to obtain the noise-corrupted edge token, that is $\texttt{ddpm\_scheduler.add\_noise}(E_j, \boldsymbol{\epsilon}, t) \rightarrow \Tilde{E}_{j}$ with $\boldsymbol{\epsilon} \sim \mathcal{N}(\mathbf{0}, \mathbf{I})$. Finally, we compute the edge token embedding as:
\begin{equation}
    \mathbf{x}_j = \varphi(\Tilde{E}_j) + \mathbf{Emb}_P(i) + \mathbf{Emb}_E(j) + \mathcal{T}(\mathbf{t}),
\end{equation}
where $\varphi(\cdot)$ and $\mathcal{T}(\cdot)$ are the projections, each comprising two linear layers with a non-linear activation function, to match the model's dimension.

\paragraph{Conditional training.}
To achieve both text-to-pattern and image-to-pattern generation with fine-grained control, we inject the conditions using cross-attention layers rather than adaLN-Zero layers~\cite{peebles2023scalable}. Specifically, we follow the practice~\cite{ye2023ip} to employ decoupled cross-attention layers that use separate key and value projection matrices to process text and image features, while using shared query projection matrices among different modalities to process edge features. The multimodal cross-attention operation is defined as:
\begin{align}
    \mathbf{Z}' = \text{Softmax}\left(\frac{\mathbf{Q}\mathbf{K}_T^\top}{\sqrt{C}}\right)\mathbf{V}_T + \text{Softmax}\left(\frac{\mathbf{Q}\mathbf{K}_I^\top}{\sqrt{C}}\right)\mathbf{V}_I,
\end{align}
where $\mathbf{Z}'$ is the fused multimodal latent features; $\mathbf{Q}$ is the edge features from $\mathbf{Z}$ after query projection; $\mathbf{K}_T$, $\mathbf{V}_T$, $\mathbf{K}_I$, $\mathbf{V}_I$ are the text and image features after key and value projection.
The text and image features are extracted using CLIP's text and image encoders~\cite{radford2021learning}. We use both class and patch embeddings before the last projection layers of CLIP for the conditional training.
The image features are projected into the same dimension as the text features by a two-layer MLP before fed into the diffusion transformer. All the parameters of the diffusion transformer are trainable. The training objective of our model is a simple $L2$ loss, which minimizes the mean-squared error between the sampled Gaussian noise and the predicted noise. That is:
\begin{equation}
    \mathcal{L}(\theta) = \mathbb{E}_{\mathbf{x}, \mathbf{c}_T, \mathbf{c}_I, \boldsymbol{\epsilon} \sim \mathcal{N}(\mathbf{0}, \mathbf{I}), t} \left[ || \boldsymbol{\epsilon} - \boldsymbol{\epsilon}_\theta (\mathbf{x}_t, \mathbf{c}_T, \mathbf{c}_I, t) ||_2^2 \right],
\end{equation}
where $\mathbf{x}_t$ is the edge token embeddings at time step $t$; $\mathbf{c}_T$ and $\mathbf{c}_I$ are the text and image conditions, respectively.

\section{Experiments}

\begin{table*}[t]
    \centering
    \small
    \begin{tabular}{p{3.5cm}lccccccr}
        \toprule
        \textbf{Method \& dataset} & $Panel L2 \downarrow$ & $\#Panel \uparrow$ & $\#Edge \uparrow$ & $Rot L2 \downarrow$ & $Trans L2 \downarrow$ & $Precision \uparrow$ & $Recall \uparrow$ & $F1 \uparrow$ \\
        \midrule
        SewingGPT \& original & 1.02e1 & 0.754 & 0.887 & 7.51e-3 & 1.25e0 & 0.833 & 0.833 & 0.825 \\
        SewingGPT \& brief & 8.89e0 & 0.936 & 0.951 & 8.51e-3 & 9.85e-1 & 0.933 & 0.933 & 0.933 \\
        SewingGPT \& detailed & 8.35e0 & 0.979 & 0.946 & 8.34e-3 & 8.93e-1 & 0.973 & 0.974 & 0.973 \\
        \midrule
        Ours \& original & 8.24e0 & 0.794 & 0.969 & \textbf{9.21e-5} & 9.42e-1 & 0.856 & 0.857 & 0.849 \\
        Ours \& brief & 7.48e0 & 0.955 & 0.999 & 1.99e-4 & 7.79e-1 & 0.956 & 0.955 & 0.955 \\
        Ours \& detailed & \textbf{6.53e0} & \textbf{0.989} & \textbf{0.999} & 2.79e-4 & \textbf{7.30e-1} & \textbf{0.989} & \textbf{0.989} & \textbf{0.989} \\
        \bottomrule
    \end{tabular}
    \caption{\textbf{Quantitative evaluation results on the DressCodeData (test set).} The metrics are explained in Section~\ref{sec:eval_metrics}. The $L2$ metrics are measured in centimeters (except $RotL2$). The first three rows show SewingGPT's evaluation results using its original captions (generated by GPT-4V) and ours (brief and detailed descriptions). The last three rows show GarmentDiffusion's evaluation results with the same text conditions. Our model outperforms SewingGPT by a large margin on different levels of captions and captions generated by different pipelines.}
    \label{tab:dresscode}
\end{table*}

\begin{figure}[t]
    \centering
    \includegraphics[width=\linewidth]{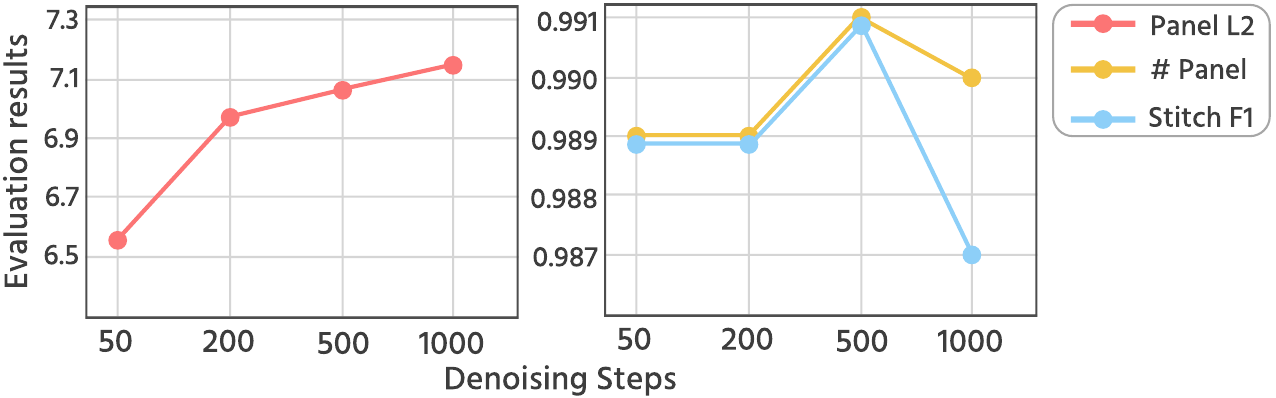}
    \caption{\textbf{Comparison of model performance with different denoising steps on the DressCodeData.} Increasing the number of denoising steps does not lead to improved model performance.}
    \label{fig:line}
\end{figure}

\subsection{Datasets}
\label{sec:exp_dataset}
We use SewFactory~\cite{Liu_2023}, DressCodeData~\cite{korosteleva2021generatingdatasets3dgarments,he2024dresscode} and GarmentCodeData (\textbf{V2})~\cite{korosteleva2024garmentcodedatadataset3dmadetomeasure} for training and evaluation. For SewFactory, we employ off-the-shelf rendered garments superimposed on diverse human poses as image prompts (without text prompts). For DressCodeData and GarmentCodeData, we designed multimodal data annotation pipelines (depicted in Figure~\ref{fig:data}) to generate both text and image prompts for sewing patterns. 
SewFactory consists of 13,707 sewing patterns, featuring a maximum of 14 panels and 12 edges per panel. DressCodeData contains 19,683 patterns, each with up to 10 panels and 10 edges per panel. GarmentCodeData offers 115,195 patterns, with a maximum of 37 panels and 39 edges per panel.
Since the official splits of SewFactory are not available, we use our own version that $90\%$ of randomly selected data points are used for training, with the remaining $10\%$ evenly divided for validation and testing. For DressCodeData and GarmentCodeData (V2), we adhere strictly to the official splits provided by the authors for training, validation, and testing. 
Note that SewingGPT cannot be trained on the entire GarmentCodeData due to its long context length. To address this problem, we created a subset of GarmentCodeData by filtering out those patterns with \texttt{\#edges/panel > 12} and \texttt{\#panels/pattern > 10}.

\subsection{Multimodal Data Synthesis}
\paragraph{Text-prompt generation.}

To enhance our model's comprehension of multi-level design concepts from users, we design two-level text descriptions for the sewing patterns: a concise category-level summary with basic design features, and detailed component-level descriptions. Leveraging LLMs' in-context understanding, our annotation pipeline first filters irrelevant information from garment specification files, then prompts Llama-3.1-8B-Instruct~\cite{grattafiori2024llama3herdmodels} using a unified prompt to generate the multi-level descriptions automatically, as shown in Figure \ref{fig:data}. 

\paragraph{Image-prompt generation.}
We select the garment sketches as the interactive interface between users and our model. To generate the garment sketches that closely emulate the hand-drawn style of professional garment designers, we initially render the 3D garment models (in \texttt{.obj} and \texttt{.ply} formats) of DressCodeData and GramentCodeData into 2D garment images using Blender's APIs~\cite{BlenderDocumentation}. Subsequently, we utilize MistoLine~\cite{zhang2023adding} and Anything-XL fine-tuned from SD-XL~\cite{podell2023sdxlimprovinglatentdiffusion} to extract the garment sketches, followed by a binarization operation.

\subsection{Evaluation Metrics}
\label{sec:eval_metrics}
We adopt the same evaluation metrics as ~\cite{Liu_2023} and ~\cite{korosteleva2022neuraltailor} to assess the fidelity of the generated sewing patterns. 
Specifically, \textbf{Panel L2} denotes the $L2$ distance of the coordinates of 2D panels (converted from 3D coordinates) between predictions and ground truths, with the centroids of panels shifted to the origin. 
\textbf{\#Panel} and \textbf{\#Edge} represent the accuracy of correctly predicted patterns within all patterns, based on the number of panels in each pattern and the number of edges in each panel, respectively.
\textbf{Rot L2} and \textbf{Trans L2} represent the $L2$ distances of $x,y,z$ rotation Euler angles and universal $x,y,z$ translations of panels between predictions and ground truths. 
\textbf{Precision}, \textbf{Recall}, and \textbf{F1 Score} are used to measure the false positives and false negatives of paired stitches with respect to all edge relations.

\subsection{Implemention Details}

\begin{table}
    \centering
    \scriptsize
    \resizebox{.48\textwidth}{!}{
        \begin{tabular}{l@{\hspace{4pt}}c@{\hspace{4pt}}c@{\hspace{4pt}}c@{\hspace{4pt}}c@{\hspace{4pt}}c@{\hspace{4pt}}r@{\hspace{3pt}}}
            \toprule
            \textbf{Input} & $P L2 \downarrow$ & $\#Panel \uparrow$ & $\#Edge \uparrow$ & $Rot L2 \downarrow$ & $Trs L2 \downarrow$ & $F1 \uparrow$ \\
            \midrule
            SewingGPT \& brief & 1.52e1 & 0.708 & 0.686 & 1.52e-2 & 2.71e0 & 0.529 \\
            SewingGPT \& detailed & 1.34e1 & 0.762 & 0.733 & 1.52e-2 & 2.03e0 & \textbf{0.589} \\
            \midrule
            Ours \& brief & 1.37e1 & 0.738 & 0.706 & 2.26e-3 & 1.90e0 & 0.463 \\
            Ours \& detailed & \textbf{1.19e1} & \textbf{0.815} & \textbf{0.786} & \textbf{1.50e-3} & \textbf{1.74e0} & 0.553 \\
            \bottomrule
        \end{tabular}
    }
    \caption{\textbf{Quantitative evaluation results on the subset of GarmentCodeData.} 
    Due to the high memory requirements for training caused by excessively long sequences in SewingGPT, 
    we filtered the GarmentCodeData to retain only sewing patterns with no more than 10 panels 
    and a maximum of 12 edges per panel (consistent with SewingGPT) for training.}
    \label{tab:dresscode-garmentcode}
\end{table}

\begin{table}
    \centering
    \scriptsize
    \begin{tabular}{l@{\hspace{4pt}}c@{\hspace{4pt}}c@{\hspace{4pt}}c@{\hspace{4pt}}c@{\hspace{4pt}}c@{\hspace{4pt}}r@{\hspace{3pt}}}
        \toprule
        \textbf{Input} & $P L2 \downarrow$ & $\#Panel \uparrow$ & $\#Edge \uparrow$ & $Rot L2 \downarrow$ & $Trs L2 \downarrow$ & $F1 \uparrow$ \\
        \midrule
        SewFormer$^{\dagger}$ & 3.76e0 & 0.859 & 0.956 & 2.01e-2 & 6.10e-1 & \textbf{0.946} \\
        \midrule
        Ours & \textbf{3.73e0} & \textbf{0.883} & \textbf{0.978} & \textbf{3.51e-3} & \textbf{6.03e-1} & 0.942 \\
        \bottomrule
    \end{tabular}
    \caption{\textbf{Quantitative evaluation results on the SewFactory (test set).} Note that the official training split of SewFactory is not provided by authors. Therefore, SewFormer$^{\dagger}$ is retrained by us using its official codebase with our split and evaluated on the same test set.}
    \label{tab:SewFormer}
\end{table}

\paragraph{Architecture.}
We adopt \textit{clip-vit-large-patch14-336}~\cite{radford2021learning} as our text and image encoders. Since the embedding dimensions of the text and image features are 1,024 and 768, we project the image features into 768-dimensional vectors to match the text feature dimension.
The main body of our model consists of $12$ DiT blocks. Each block consists of a self-attention layer, a multimodal cross-attention layer and a feed-forward layer, all utilizing pre-layer normalization~\cite{ba2016layernormalization,xiong2020layer}. The number of heads for each attention layer is set to $8$. The embedding dimension of the DiT blocks is $768$, while the feed-forward layers have an embedding dimension of $1,024$.

\paragraph{Look-up embedding tables.}
The panel-level and edge-level embedding tables contain $M$ and $N$ learnable positional embeddings, where $M$ is the maximum number of panels and $N$ is the maximum number of edges per panel in each dataset. For SewFactory, $M=14$, $N=12$. For DressCodeData, $M=N=10$. For GarmentCodeData, $M=37$, $N=39$.

\paragraph{Training details.}
We adopt a DDPM noise scheduler for diffusion training, with a maximum of $1,000$ denoising steps and a linear beta scheduler ($\text{beta\_start}=1 \times 10^{-4}, \text{beta\_end}=2 \times 10^{-2}$). We use the AdamW optimizer~\cite{loshchilov2019decoupledweightdecayregularization} with $\text{betas}=(0.95,0.999)$, a constant learning rate of $1 \times 10^{-4}$ and the weight decay of $1 \times 10^{-2}$. The training epoch is set to $1,000$ with an early-stop criterion. 
We evaluate the model at denoising steps of 50, 200, 500, and 1000 every 10 epochs. Based on the results shown in Figure \ref{fig:line}, we select 50 denoising steps for inference.
The multimodal training is performed in a round-robin fashion, following the order of image prompts, text prompts and image-and-text prompts. Our model is distributedly trained across 8 A10 GPUs (24GB) with the Hugging Face Accelerate library~\cite{accelerate}.

\subsection{Comparison with State-of-the-Art Methods}
\paragraph{Compared with SewingGPT.}

Our evaluation metrics assess the accuracy of geometric structures, panel placement in 3D space, and stitching relations. 
Table~\ref{tab:dresscode} shows the evaluation results for SewingGPT and GarmentDiffusion, using text prompts from two different annotation pipelines.
The first three rows show that the text descriptions generated by GPT-4V with DressCode's pipeline result in worse performance than ours.
This is expected, as we prompt the Llama3.1-8B-Instruct (text-only)~\cite{llama3.1-8b-instruct} with precise design specifications, which are used to generate sewing patterns through programs.
Moreover, our method outperforms SewingGPT when trained with the same captions, highlighting the advantage of our approach.

\begin{figure}[t]
    \centering
    \includegraphics[width=\linewidth]{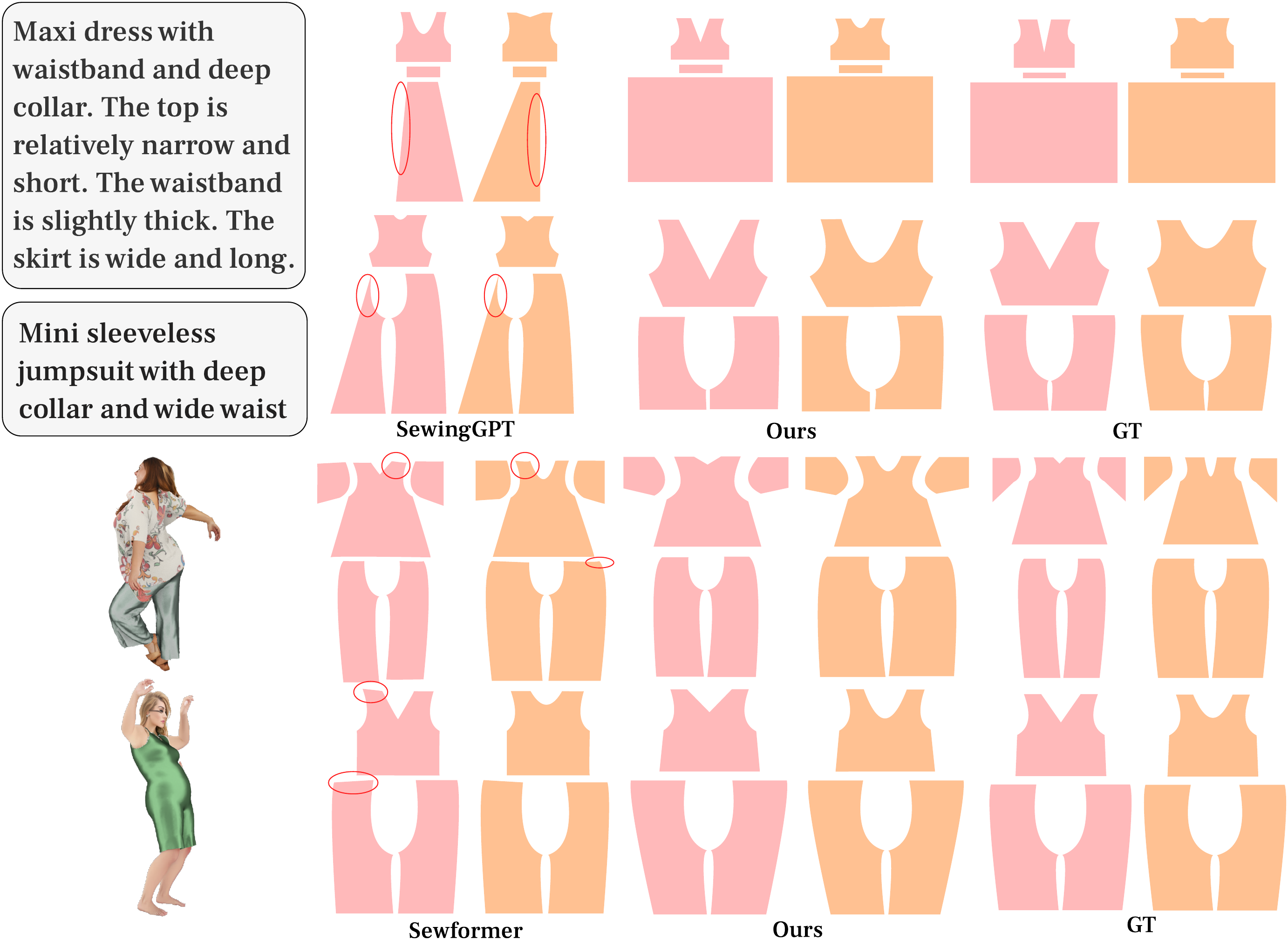}
    \caption{\textbf{Visualization of patterns generated with SewingGPT, SewFormer and ours.}
    Major errors of the baseline approaches are highlighted with red circles.}
    \label{fig:compare}
\end{figure}

We also evaluated SewingGPT on the subset of GarmentCodeData. 
As shown in Table \ref{tab:dresscode-garmentcode}, 
our method outperforms SewingGPT in panel and edge number accuracies, geometrical shapes, and 3D placement of panels.
However, it lags in stitching edge prediction, likely due to the lack of stitching information in the prompt.

\begin{figure}[t]
    \centering
    \includegraphics[width=\linewidth]{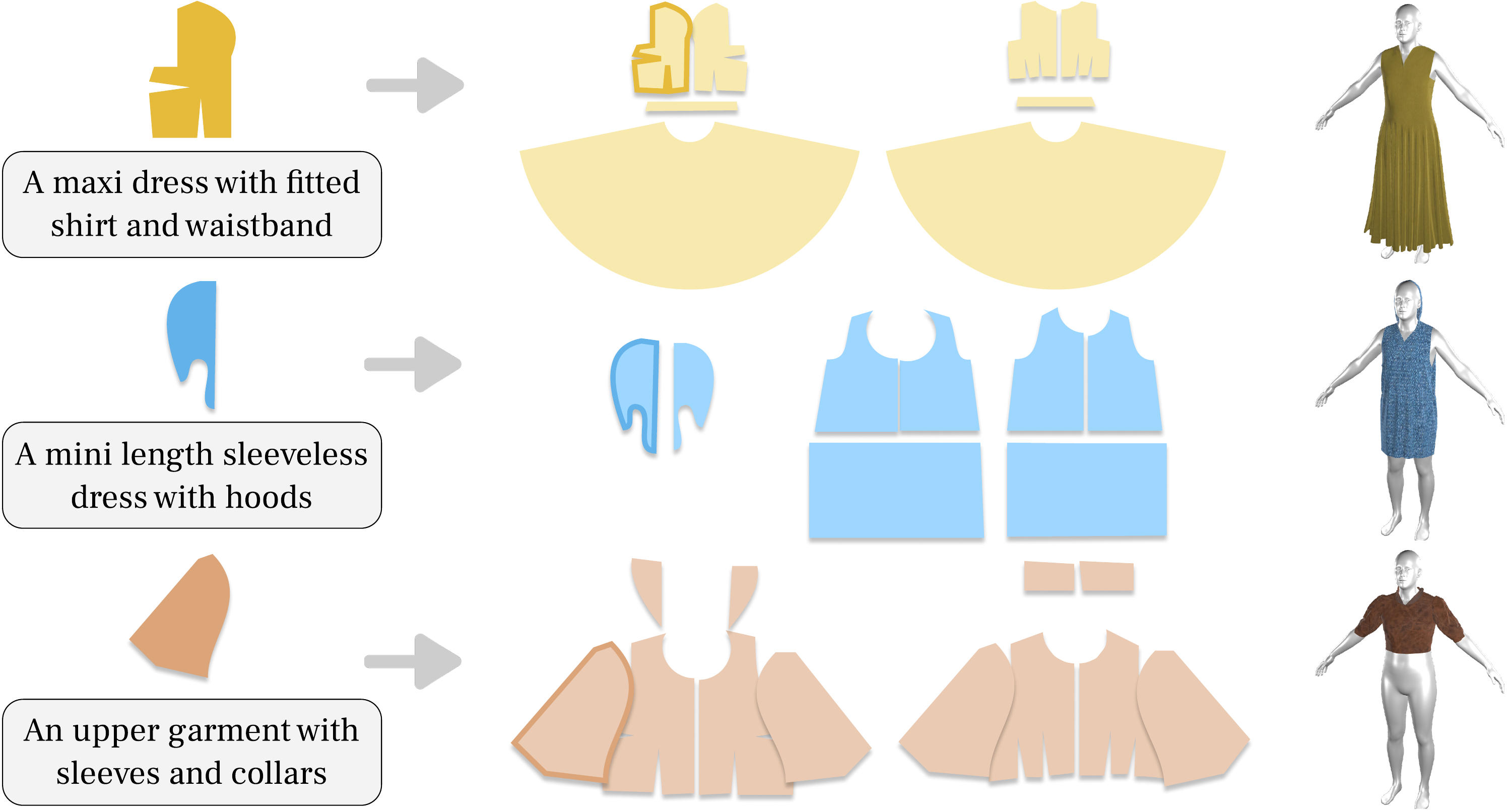}
    \caption{\textbf{Visualization of pattern completion.}
    Given a ground-truth panel and a text prompt, our model is capable of generating a consistent and complete sewing pattern.}
    \label{fig:autocomplete}
\end{figure}

As shown in Figure \ref{fig:autocomplete}, our model also supports pattern completion with incomplete patterns provided by users. It achieves strong control even though the text descriptions are not precise.

\paragraph{Compared with SewFormer.}
We also trained SewFormer and GarmentDiffusion using our training split as described in Section~\ref{sec:exp_dataset}, while maintaining the same evaluation protocols and test set to ensure fair comparison. Table \ref{tab:SewFormer} demonstrates that our generative model achieves comparable performance to SewFormer in terms of the geometrical shapes and 3D panel placement, while it surpasses SewFormer in terms of panel and edge accuracies. These results confirm the effectiveness of our method, even when using multi-pose rendered images as prompts.

\subsection{Ablation Study}

\begin{table*}[t]
    \centering
    \small
        \resizebox{1.0\textwidth}{!}{
            \begin{tabular}{lccccccccccr}
            \toprule
            \textbf{Train Modality} & \multicolumn{2}{c}{\textbf{Text}} & \multicolumn{1}{c}{\textbf{Image}} & $Panel L2 \downarrow$ & $\#Panel \uparrow$ & $\#Edge \uparrow$ & $Rot L2 \downarrow$ & $Trans L2 \downarrow$ & $Precision \uparrow$ & $Recall \uparrow$ & $F1 \uparrow$ \\
            & brief & detailed & sketch & \\
            \midrule
            MM (text\&image) & \checkmark & & & 1.31e1 & 0.388 & 0.600 & 2.26e-3 & 1.69e0 & 0.380 & 0.364 & 0.364 \\
            MM (text\&image) & & \checkmark & & 1.12e1 & 0.464 & 0.701 & 2.07e-3 & 1.53e0 & 0.472 & 0.457 & 0.460 \\
            MM (text\&image) & & & \checkmark & 1.08e1 & 0.537 & 0.713 & \textbf{1.70e-3} & 1.38e0 & 0.430 & 0.431 & 0.425\\
            MM (text\&image) & \checkmark & & \checkmark & 7.48e0 & 0.616 & 0.771 & 1.90e-3 & 1.00e0 & 0.516 & 0.506 & 0.509\\
            MM (text\&image) & & \checkmark & \checkmark & \textbf{6.68e0} & \textbf{0.670} & \textbf{0.819} & 1.85e-3 & \textbf{9.26e-1} & \textbf{0.564} & \textbf{0.561} & \textbf{0.560} \\
            \midrule
            Text-only & \checkmark & & & 1.52e1 & 0.287 & 0.470 & 3.13e-3 & 2.00e0 & 0.270 & 0.239 & 0.244\\
            Text-only & & \checkmark & & 1.08e1 & 0.486 & 0.707 & 2.60e-3 & 1.61e0 & 0.474 & 0.473 & 0.469\\
            \midrule
            Image-only & & & \checkmark & 1.05e1 & 0.528 & 0.723 & 1.91e-3 & 1.31e0 & 0.443 & 0.436 & 0.435\\
            \bottomrule
            \end{tabular}
        }
    \caption{\textbf{Quantitative evaluation results of GarmentDiffusion on the GarmentCodeData (test set).}
    The first five rows are trained using both text and image prompts and evaluated with different modality combinations. 
    The two rows in the middle are trained with text prompts only and evaluated with brief or detailed text descriptions. The last row is trained using image prompts. We use the whole GarmentCodeData for training.}
    \label{tab:garmentcode-mm}
\end{table*}

\paragraph{Inputs of different modalities.}
To assess the impact of training on model performance under different combinations of input modalities, we trained three models using text, image, and multimodal prompts on the GarmentCodeData, respectively. As shown in Table~\ref{tab:garmentcode-mm}, under the multimodal training setting, evaluation metrics are gradually improved when fine-grained conditions are incorporated for generation. Specifically, the combination of detailed text descriptions and sketches achieve the best performance. This conclusion is also supported by the middle section of the table. Compared with the model trained using text-only prompts, the model trained with image-only prompts exhibits slightly better performance. This is reasonable, as images may convey more information than texts.

\paragraph{Condition injection method.}
In addition to employing cross-attention for conditional training, 
we also experimented with adaLN-Zero conditioning proposed in DiT~\shortcite{podell2023sdxlimprovinglatentdiffusion}. 
We trained multimodal models (text \& image) with both conditioning methods on the augmented DressCodeData. 
For cross-attention, we evaluate the model under text, image, and text-and-image conditions, 
while for adaLN-Zero, we used text and image conditions with equal probability. 
As shown in Table \ref{tab:adaLN-Zero}, although adaLN-Zero exhibits advantages when evaluated with brief text descriptions, 
it becomes less effective with detailed text descriptions and image inputs due to insufficient conditional information extraction. 
Cross-attention achieves the best overall performance when both detailed text and image inputs are provided.

\begin{table}
    \centering
    \scriptsize
        \resizebox{.48\textwidth}{!}{
            \begin{tabular}{lcccc@{\hspace{3pt}}c@{\hspace{3pt}}r@{\hspace{3pt}}}
            \toprule
            \textbf{Condition Scheme} & \multicolumn{2}{c}{\textbf{Text}} & \multicolumn{1}{c}{\textbf{Image}} & $Panel L2 \downarrow$ & $\#Panel \uparrow$ & $F1 \uparrow$ \\
            & brief & detailed & sketch & \\
            \midrule
             & \checkmark & & & 7.42e0 & 0.944 & 0.949  \\
             & & \checkmark & & 6.64e0 & 0.994 & 0.995  \\
             Cross-attention & & & \checkmark & 2.47e0 & 0.978 & 0.985  \\
             & \checkmark & & \checkmark & 2.41e0 & 0.994 & 0.994  \\
             & & \checkmark & \checkmark & \textbf{2.39e0} & \textbf{0.994} & \textbf{0.995}  \\
            \midrule
             & \checkmark & & & 7.49e0 & 0.958 & 0.949  \\
            AdaLN-Zero & & \checkmark & & 7.54e0 & 0.970 & 0.959  \\
             & & & \checkmark & 3.25e0 & 0.968 & 0.979  \\
            \bottomrule
            \end{tabular}
        }
    \caption{\textbf{Quantitative evaluation results for different condition injection methods.}
    The first five rows present the evaluation results using the cross-attention method, while the last three rows correspond to the conditioning using the adaLN-Zero method. Both methods utilizes text and image prompts for training. We report three metrics to save the space in the table.}
    \label{tab:adaLN-Zero}
\end{table}

\subsection{Limitations and Future Works}

While current annotations provide detailed sewing pattern descriptions, they still lack stitching information on edge and panel connectivity, which can compromise garment simulation. The annotation engine thus remains improvable. Additionally, current methods offer limited control via numerical parameters (e.g., panel/edge count) or human body measurements. From an efficiency standpoint, reducing denoising steps is also desirable. Future work will target these challenges by enhancing controllability and generation efficiency.

\section{Conclusion}
In conclusion, we introduced GarmentDiffusion, a much under-explored research direction for sewing pattern generation. Our model incorporates the design of diffusion transformers with an efficient edge encoding scheme. The architecture and training of our model are simple yet efficient, enabling the end-to-end generation of centimeter-precise and vectorized 3D sewing patterns. Our experimental results demonstrate the effectiveness of our method, which bridges the gap between creative garment design and manufacturing through scalable, precise, and efficient generative modeling. This work lays the foundation for advancing AI-driven fashion technology, seamlessly connecting digital design with practical garment production.

\bibliographystyle{named}
\bibliography{ijcai25}

\end{document}